# High Performance Software in Multidimensional Reduction Methods for Image Processing with Application to Ancient Manuscripts

Corneliu T.C. Arsene*[1], Stephen Church[2], Mark Dickinson[2]

[1]School of Arts, Languages and Cultures, University of Manchester, United Kingdom
[2]Photon Science Institute, University of Manchester, United Kingdom
email: corneliu.arsene@manchester.ac.uk, arsenecorneliu@yahoo.co.uk,
stephen.church@manchester.ac.uk, mark.dickinson@manchester.ac.uk

Abstract

Multispectral imaging is an important technique for improving the readability of written or printed text where the letters have faded, either due to deliberate erasing or the ravages of time. Often the text can be read by illumination under a single wavelength of light, but in some cases the multispectral images need enhancement to improve the text clarity. There are many possible enhancement techniques: this paper compares an extended set of dimensionality reduction methods for image processing. We assess 15 dimensionality reduction methods applied to two different manuscripts. This assessment was performed subjectively, by asking the opinions of scholars who were experts in the languages used in the manuscripts, and also by using the Davies-Bouldin and Dunn indexes for evaluating the quality of the resultant image clusters. We found that the Canonical Variates Analysis (CVA) method, implemented in Matlab was superior to all the other tested methods. However, the other approaches may be more suitable in specific circumstances, so we would still recommend that a variety are tried. For example, CVA is a supervised clustering technique and therefore it requires considerably more user time and effort than a non-supervised technique such as the Principle Component Analysis approach (PCA). If the results from PCA are adequate to allow a text to be read then the added effort required for CVA may not be justified. For the purposes of comparing the computational times and the image results, a CVA method is also implemented in the C programming language and using the *GNU* (*G*NU's *N*ot *U*nix) *S*cientific *L*ibrary (*GSL*) and the OpenCV (OPEN source Computer Vision) computer vision programming library. Therefore high performance software was developed using the GNU GSL library, which drastically reduced the computational complexity and time for the CVA-GNU GSL method. For the CVA-Matlab technique, vectorization was used in order to reduce the respective computational times (i.e. matrix and vector operations instead of loop-based).

## 1. Introduction

Multispectral/hyperspectral image analysis has undergone much development in the last decade[1]: it has become a popular technique for imaging hard-to-read documents[2], as it is a non-invasive and non-destructive way of analyzing such documents by fully utilizing the full light spectrum. Multispectral images are obtained by illuminating the document or manuscript with narrow band light sources at a set of different wavelengths ranging from ultraviolet (300 nm) to infrared (1000 nm). This technique was applied to *Archimedes Palimpsest*[2], which was imaged over several years and in several stages with different imaging systems as technology was developed. A palimpsest is an old manuscript for which the initial writing (i.e. the underwriting) was deleted and new writing (i.e. the overwriting) was written on the same parchment. In the most recent work, 16 spectral images were selected, which included one waveband centered in the near Ultra Violet (UV) region of the spectrum (365 nm), seven visible bands (445 nm, 470 nm, 505 nm, 530 nm, 570 nm, 617 nm and 625 nm) and three infrared bands (700 nm, 735 nm, 870 nm). In addition, images were obtained under illumination with a tungsten lamp and under raking illumination using separate lighting units from two sides and at two wavelengths (470 nm and 870 nm). The inks used in the writing of the manuscript such as iron-gall inks, have a relatively low reflectance at UV wavelengths[2]. Since the parchment is highly reflective in the UV, the ink can appear dark against a brighter background, which has a high reflectance at UV wavelengths. This distinction would be lost in conventional images, which are formed by integrating over larger wavelength ranges, and any variations in wavelength dependent reflectance or absorption, tend to average out.

---

[1] Kwon et al. 2013; Wang and Chunhui 2015; Shanmugam and SrinivasaPerumal 2014; Chang 2013; Zhang and Du 2012.

[2] Easton, Christens-Barry and Knox 2011; Easton and Noel 2010; Netz et al. 2010; Easton et al. 2010; Bermann and Knox 2009, URL: http://www.archimedespalimpsest.net.

The application of these multispectral/hyperspectral image analysis methods to old manuscripts and palimpsests[3] has produced significant results, enabling researchers to recover texts that would be otherwise lost, by improving the contrast between the text and the manuscript.

A number of important old manuscripts and ancient palimpsests have been processed in the last decade with significant results by using not only multispectral imaging systems and enhanced image processing techniques but also synchrotron facilities[4]. The *Archimedes Palimpsest*[5] was processed between 2000 and 2011 with great success. The *Archimedes Palimpsest* is a circa 10$^{th}$ century parchment manuscript that was deleted in the early of the 13$^{th}$ century and overwritten with a Christian prayer book called the *Euchologion*. The palimpsest was named as such because in the early of the 20$^{th}$ century it was identified to be partial copies of seven scientific documents by *Archimedes*, the oldest surviving reproductions of his writings. During the respective project, an extended number of multispectral imaging techniques and processing methods were developed and successfully applied on the respective palimpsest. Multispectral imaging techniques[6] and image processing were carried out for manuscripts originating from Mount Sinai in Egypt, which were written between the 10$^{th}$ and 12$^{th}$ centuries in Glagolitic, the oldest Slavonic script. A number of palimpsests[7] which originated from the New Finds at Saint Catherine's Monastery in Sinai in today's Egypt were also analysed using multispectral imaging techniques. A number of old manuscripts or documents[8] (e.g. paintings) were multispectral imaged and processed namely the documentation of Heinrich Schliemann's (1822-1890) copybooks, Nikolas Gyzis's oil sketches, who was an important Greek painter of the 19$^{th}$ century, and an old papyrus dated 420/430 BC, which is the Oldest Greek text, discovered in Daphne, Greece. In the San Lorenzo palimpsest[9], multispectral imaging revealed underwriting, which contained over 200 secular musical compositions dated from the 14$^{th}$ and the beginning of the 15$^{th}$ century. The overwriting contained the church properties until the 17$^{th}$ century in Florence in Italy (i.e. document named *Campione dei Beni*). It is worth mentioning that other methods have been used to study old manuscripts. For example, X-ray fluorescence analysis[10] was performed to discover the history of making of the *Codex Germanicus 6*, which is a combination of twelve different texts forming a 614-page manuscript created around 1450 in Germany. The X-ray Phase-Contrast Tomography (XPCT) technique[11] was used to uncover letters and words in two papyrus rolls, which were buried by the eruption of Mount Vesuvius in 79 AD, belonging at that time to a library in Herculaneum, Italy. A combination of X-ray fluorescence and multispectral imaging[12] was applied to the study of Leonardo da Vinci's *The Adoration of the Magi* drawing. Synchrotron radiation X-Ray Fluorescence (srXRF)[13] has been used also in the study of other old manuscripts.

The image processing techniques discussed in this paper have been applied, for the purposes of image enhancement, to two old manuscripts. The first manuscript is with regard to Aelius Galenus (ca. 129-216), who was an important Greek physician and philosopher in the Roman empire and influenced the development of various scientific disciplines such as anatomy, physiology, pathology and pharmacology. For many centuries, the book written by Galen *On Simple Drugs*, was required to be known while seeking to become a physician as the book contained ancient knowledge about pharmaceutical plants and medicine. The Syriac Galen palimpsest[14] is an important ancient manuscript, which put many challenges to researchers as the undertext contains the Syriac translation[15] by Sergius of Reshaina of Galen's *On Simple Drugs*. Sergius of Reshaina was a Christian physician and priest (d.536). This palimpsest is especially important because it contains more text than in the other historical copy of the Syriac translation made by Sergiu of Reshaina, which exists in London, British Library (BL), MS Add. 14661. There are also better readings than in the other historical copy existent in BL. The text has relevance to both the Greek source text, the Arabic target texts and the development of Greco-Arabic translation technique. Finally, it is able to address the role of Sergius' Syriac versions on Hunayn ibn Ishaq's school[16].

---

[3] Bhayro, Pormann and Sellers 2013; Pormann 2015; Hollaus, Gau and Sablatnig 2012.

[4] Rabin, Hahn and Geissbuhler 2014; Mocella et al. 2015.

[5] Walvoord and Easton 2008; Easton, Christens-Barry and Knox 2011; Easton and Noel 2010; Netz et al. 2011; Easton et al. 2010; Bergmann and Knox 2009.

[6] Camba et al. 2014.

[7] Easton and Kelbe 2014.

[8] Alexopoulou and Kaminari 2014.

[9] Janke and MacDonald 2014.

[10] Rabin, Hahn and Geissbuhler 2014.

[11] Mocella et al. 2015.

[12] Stout, Kuester and Seracini 2012.

[13] Glaser and Deckers 2014; Manning et al. 2013.

The second manuscript on which image processing techniques will be applied is an old Latin Roll titled *John Rylands Library, University of Manchester, United Kingdom, Latin MS 18*, with its catalogue entry dating from 1921[17] and entitled *Arbor Caritas at Misericordiae*. Although this Roll does not have any underwriting (i.e not a palimpsest), some of the text has been almost deleted on account of water damage and the effects of age. This Roll is also peculiar as it contains many illuminations/drawings, such as Church Fathers and saints, and biblical images from the Old and New Testaments.

## 2. Dimensionality Reduction Methods

There are various ways to improve the quality of an image of a page of a manuscript such as deblurring, enhancement or dimensional reduction methods.

Deblurring[18] of images involves of removing out blurring items from images which are caused, for example, by the fact that the image is out of focus. Deblurring can be done in several ways, such as by using a Wiener filter, a regularized filter, a blind deconvolution algorithm or the Lucy-Richardson algorithm[19]. Image enhancement can be achieved by modifying the histogram of pixel values in the image to adjust the contrast, thereby improving the clarity of details within the picture. This is typically achieved by linearly scaling the pixel values between two reference points, however, in some cases significant improvements can be made by using polynomial scaling to higher orders, ie. $L^2$, $L^3$ or $L^4$, which can place more emphasis on the variation within features of interest. Color images can be also enhanced by transforming the RGB images to L*a*b* color space and then by altering the luminosity L* of the image. Techniques to improve the image contrast between the manuscript and the written text were developed elsewhere[20] and included, for example, the implementation of a custom image look-up table to display the text in false-color, the automatic contrast adjustment of the image based upon the quartic scaling of pixel values and the removal of variation in the manuscript pixels by blurring out the image details and subtracting the original image. These methods allowed an inexperienced user to maximize the clarity of text, but were heavily dependent upon required sampling and could produce artifacts and incoherent results due to large scale image variation and sampling mistakes. Therefore additional methods were developed to carry out the study with no user input. This included the calculation of localized variances, which provide a distinct outline for any text based upon the large change in pixel value between the text and other image components, and 2 dimensional spatial autocorrelation indexes, which distinguished between text and the manuscript based on the degree of variation in each region.

---

[14] Bhayro et al. 2013, URL: http://www.digitalgalen.net.

[15] Montgomery 2000; Montgomery 2001.

[16] Khurshid 1996.

[17] James 1921.

[18] Shao and Elad, 2015.

[19] Fish et al. 1995.

[20] Church 2015.

Color image enhancement in multispectral images can also be achieved by using the Karhunen-Loeve transform[21], a linear contrast stretch or a decorrelation stretch. Moreover, contrast enhancement techniques based on histograms for multispectral images have been developed in the past[22]. This method was further developed and applied to multispectral images[23] to enhance the images which are not in the visible range. Multispectral image enhancement techniques based on PCA and Intensity-Hue-Saturation (IHS) transformations have also been developed and applied[24].

For palimpsests such as the *Archimedes Palimpsest*, image enhancing techniques have been applied to the acquired multispectral images[25]. Initially, spectral segmentation techniques were developed[25] based on a least-squares supervised classification, but the scholars assessed that the results were not clear enough. Following this, the contrast of each image band was enhanced using the neighborhood information of a pixel and then subtracting two resultant channels (i.e. called sharpies method). However, the subtraction increased the noise, so a new method called "pseudocolor" was developed. In this technique, the red channel under tungsten illumination was placed in the red channel of a new composite image, while the ultraviolet-blue image was placed in the blue and green channel. In this way, the overwriting appeared gray while the underwriting was red in the composite image. Finally, the PCA method was employed, which provided further enhancement to the multispectral images. The application of a combination of the PCA method with the pseudocolor method provided the best quality in the investigations.

An example of a regular and simple image enhancement technique for image processing briefly used herein is a Double Thresholding (DT) technique, which consists of the following: the darker overtext is carefully identified by the human operator and colored in white (threshold 1), and then the remaining undertext, which is slightly lighter than the initial overtext is made darker (threshold 2). This technique showed some initial interesting results but its success depends on both the human operator, who has to select suitable cutting values, and on the characteristics of the respective image. Therefore, this simpler method would not work for any page of an ancient manuscript. However, these image processing methods, although able to provide workable images of undertext, for example in the gutter region of a folio, are unable to show when there is undertext beneath the overtext. More complex methods[26] have been developed and are available for image processing (i.e. image reconstruction, image restoration, image segmentation) based, for example, on Artificial Neural Networks[27] (ANNs). These are information processing models, which try to mimic the way the brain works.

One solution is to use dimensionality reduction methods, which reduce the number of features or random variables under consideration by transforming the data to a space of a fewer dimensions[28]. There are two types of dimensionality reduction methods: unsupervised methods, which use a number of points to determine the model without knowing the classes (e.g. parchment, overwriting, underwriting) to which the input data points belong to, and supervised methods, which use a number of input points to determine the model while knowing the classes (e.g. parchment, overwriting, underwriting). In general, the supervised methods produce better results as they use class information and so the mathematical model is able to better reflect the sample. However, selecting a number of input points is time consuming[29], so the unsupervised methods could be advantageous, especially if an automatic method of choosing the input points could be provided.

There are a large number of dimensionality reduction methods that have been developed in the last decade and implemented in various computer programming languages. An extended number of dimensionality reduction methods have been tested in this paper by using a Matlab toolkit[30], the ones presented below are those which provided meaningful image results.

---

[21] Mitsui et al. 2005.

[22] Mlsna and Rodriguez 1995; McCollum and Clocksin 2007.

[23] Hashimoto et al. 2011.

[24] Lu et al. 2011.

[25] Easton and Noel 2010.

[26] Zhenghao et al. 2009; Doi 2007; Egmont-Petersen, de Ridder and Handels 2002.

[27] Graves et al. 2009; Lisboa et al. 2009; Arsene, Lisboa and Biganzoli 2011; Arsene and Lisboa 2011; Arsene and Lisboa 2007; Arsene et al. 2006.

[28] Freeden and Nashed 2010.

[29] Hollaus, Gau and Sablatnig 2013.

The Canonical Variates Analysis[31] (CVA) supervised method, with an independent implementation in Matlab[32], tries to maximize the distance between the different classes, whilst minimizing the size of each class. This is performed for multiple classes. The covariance matrixes within each class and between the classes are calculated and eigenanalysis is performed based on these two matrixes. The eigenvectors calculated by this eigenanalysis are the canonical vectors, which are used to produce new grayscale images. The Linear Discriminant Analysis (LDA) is similar to CVA but it is applied to 2 classes only. These methods are very robust, producing very good results, since they are both supervised. In addition, this process of maximizing the distance between the different classes while minimizing the size of each of the classes, is of key importance in the present analysis. As they are supervised methods, the human operator selects a number of points to be used and also provides the classifications of the respective points (i.e. class manuscript, class underwriting, etc).

The Neighborhood Component Analysis (NCA) supervised method, taken from the Matlab toolkit[30], is a learning algorithm for classifying multivariate data into distinct classes by using a distance metric over the data. Typically the Mahalanobis distance measure is used. The method consists of learning a linear transformation of the input space, which in this case are the multispectral images, such that in the transformed space the k-Nearest Neighbors (kNN) algorithm performs well.

From the same toolkit, the supervised General Discriminate Analysis (GDA) applies the methods of the general linear model to the discriminant function analysis problem. The advantage of doing this is that it is possible to specify complex models for the set of predictor variables continuous or categorical (e.g. polynomial regression model, response surface model, factorial regression, mixture surface regression).

Both NCA and GDA are expected to deliver some good image results as they are supervised methods and have previously been reported as effective methods in the context of multispectral/hyperspectral image analysis[33]. However, the NCA method is based on optimization algorithms (i.e. line search optimization method) for calculating the models parameters and sometimes, depending on the optimization algorithms being used, a local minimum point can be reached, which means that the image results might not be optimal. In such situations, re-running the respective dimensionality reduction method (e.g. NCA, GDA) might avoid reaching a point of local minimum.

Other dimensionality reduction methods will be used from the Matlab toolkit[30] in a supervised way, with the user selecting points as with the supervised methods, but without providing the class information.

Isomap is a nonlinear dimensionality reduction technique and, in this work, a priori chosen points are used as input information, hence this is a supervised method. Here, a matrix of shortest distances between all of the input points is constructed and multidimensional scaling is then used to calculate a reduced-dimension space. To perform the multidimensional scaling, various nonlinear methods are applied, which map the high dimensional data to a low dimensional space by attempting to maintain the original distances between points. The quality of mapping is given by a stress function, which is a measure of the error between the distances in the initial high dimensional representation and the distances in the new lower dimensional representation.

The Landmark Isomap algorithm is a variant of the Isomap, which uses landmarks to increase the speed of the algorithm. It addresses the computational load with regard to the calculations of the shortest path distances between points when reducing the dimensionality of the data and calculating the eigenvalues. A smaller number of landmark points are chosen, for which the shortest distances between the respective points and each of the other data points are calculated. This results in a reduction of the computational time. In the literature, the Isomap type of methods which are nonlinear, local and geodesic methods were reported as having good performance when tested on multispectral images[34].

The Principal Component Analysis (PCA) method is a statistical method, which performs an orthogonal transformation of the input data in order to change a number of observations of possibly correlated variables into a number of linear uncorrelated variables named Principal Components (orthogonal). The first principal component has the largest variance and therefore is responsible for most of the variation in the data. There are several stages in this well established method: subtract the mean of each variable from the dataset, calculate the covariance matrix, calculate the eigenvectors and eigenvalues of this matrix, orthogonalize the set of eigenvectors and normalize them. In the Probabilistic Principal Component Analysis (PPCA) method the principal components are calculated through maximum-likelihood estimation of parameters in a latent variable model, which offers a lower dimensional representation of the data and their correlations. The Gaussian Process Latent Variable (GPLV) model is a probabilistic dimensionality reduction method that uses a Gaussian Process Latent Variable model to find a lower dimensional space for the high dimensional data. It is an extension of the PCA. The latent variable models use, for example, one latent variable to aggregate several observable variables, which are dependent somehow (e.g. "sharing" variance). The PCA types of dimensionality reduction methods have been applied with success to more general multispectral/hyperspectral images[35] but also to old manuscripts[36].

---

[30] van der Maaten and Hinton 2008; https://lvdmaaten.github.io/drtoolbox/.

[31] Macfie, Gutteridge and Norris 1978; Maxwell 1961; Rao 1948; Fisher 1936; Campbell and Atchley 1981; Peltier, Visalli and Schlich 2015.

[32] Bohling 2010.

[33] Goldberger et al. 2005; Imani and Ghassemian 2014.

[34] Journaux et al. 2006; Journaux, Foucherot and Gouton 2008.

In this work, three variants of the PCA method will be used. One variant will be used in a supervised way by providing a set of input points chosen to represent the classes of interest (i.e. underwriting, overwriting, manuscript) and two unsupervised PCA methods implemented in ImageJ software and Matlab, which will use an entire manuscript page/folio as an input without any supervised information.

The Diffusion Maps (DM) model is a nonlinear dimensionality reduction method, which uses a diffusion distance as a measure between the different input points, and so builds a map, to provide a global description of the dataset. An analogy can be seen between a diffusion operator on a geometrical space and a Markov transition matrix operating on a graph whose nodes are sampled from the respective geometrical space (i.e. dataset). The algorithm is robust to errors and it is computationally less expensive. The DM has been applied to image processing with some success[37] and here it will be used as a supervised method.

The t-Distributed Stochastic Neighbor Embedding (t-SNE) method is another dimensionality reduction method in which the Kullback-Leibler divergence distance between two probability distributions is minimized. In the first distribution the nearby points have higher probability than the other points in the higher initial dimensional space, while the second distribution contains the lower desired dimensional space of data. The model has only been recently developed and therefore its properties are still being heavily investigated[38] in order to improve its performances and it will be used here as a supervised technique.

The Neighborhood Preserving Embedding (NPE) method tries to maintain the local neighborhood structure in the data in order to be less influenced by errors than other techniques. It is similar to the Locality Linear Embedding (LLE) method which also attempts to find a linear combination of neighbors for each input point. In addition, the LLE method implements an eigenvector-based optimization method, which is different from the one used by NPE, in order to find a low-dimensional embedding of the points in such a way that each input point is still represented by the same mixture of its neighbors. In effect, a neighborhood map is realized in the two methods (NPE, LLE), which map creates a new point in the new lower dimensional space for each point in the higher initial dimensional space. The Hessian Locally-Linear Embedding (HLLE) method is based on the LLE method in the way that it achieves a linear embedding by minimizing the Hessian functional on the data space. The HLLE algorithm involves the second derivative and therefore the algorithm is sensitive to noise.

Other methods from the same Matlab toolbox, but which resulted in poor image results, were the Factor Analysis (FA) method and the Laplacian Eigenmaps (LE) method. The FA method depicts the diversity of input data as a function of some unseen variables called factors. The observed variables are depicted as linear mixtures of the unseen factors and some error variables are also included. The information extracted with regard to the relationships between the observed variables (i.e. the correlation matrix), is used to calculate both the factors and the new reduced dimensional space by minimizing the difference between the correlation matrix of the initial input data and the correlation matrix of the new reduced space. The LE method employs spectral techniques to implement the dimensionality reduction. A graph is built in which each node represents a data point, the connections with the other graph nodes are given by the distance between such a data point and the initial data point in the higher dimensional space. The lower dimensional space is represented by the eigenfunctions of the Laplace-Beltrami operator, while the minimization of an error function based on the graph ensures that the new points in the lower dimensional space maintain the proximity characteristic of the initial data points. The calculation of the connections in the graph is difficult, which reduces the robustness of the method. The FA and LE results are not shown herein as the image results were poor.

---

[35] Journaux, Foucherot and Gouton 2008; Baronti et al. 1997; Ricotta and Avena 1997.

[36] Eastonm, Christens-Barry and Knox 2011; Easton and Noel 2010.

[37] Gepshtein and Keller 2013; Xu et al. 2009; Freeden and Nashed 2010.

[38] Bunte et al. 2012.

The above extended set of dimensionality reduction methods implemented by the Matlab toolbox[30], together with the previous results[39] obtained with the Canonical Variates Analysis (CVA) method, and the unsupervised PCA method implemented in ImageJ software and Matlab, were applied to the 102v-107r_B page of the Galen palimpsest.

Image data consisted of large 8-bit TIFF image files. There were 23 multispectral images in total, obtained through Light-Emitting Diode (LED) illumination at wavelengths of 365 nm, 450 nm, 470 nm, 505 nm, 535 nm, 570 nm, 615 nm, 630 nm, 700 nm, 735 nm, 780 nm, 870 nm, 940 nm, images obtained at raking light under illumination at 940 nm (raking infrared with illumination from the right and then from the left), 470 nm (raking blue with illumination from the right and then from the left) and ultraviolet images (365 nm) and blue illumination (450 nm) with red, green and blue color filters. The multispectral images were normalized with values between 0 and 255 and were used as an input to the all the image processing methods used in this work without any further pre-processing.

In this experiment, for all the techniques except the unsupervised PCA methods, 50 points there were selected from each of the images to represent the overwriting (Class Overwriting), the underwriting (Class Underwriting), the parchment (Class Parchment) and both overwriting and underwriting (Class Both). In the case of Class Both, the scholar could infer the presence of the underwriting from the overlapping (i.e. the structure) of the overwriting and the underwriting. There were 200-classification input points in total used by each supervised dimensionality reduction method. The input data matrix consisted of 23 rows and 200 columns. For the CVA, LDA, GDA and NCA methods information about the classes was also provided to the Matlab software. Moreover, the number of points for each class could be varied to put more emphasis on a particular class and the number of classes could also be varied, such as to exclude Class Both or to include another class representing, for example, the region outside the manuscript (Class Outside).

For the unsupervised PCA methods implemented in ImageJ and Matlab, no a priori known information was used except the entire manuscript page/folio. In ImageJ the Multivariate Statistical Analysis (MSA) 514 plugin was used, which implements the PCA method. The 23 multispectral images were loaded as a stack in ImageJ, then the Crop function from ImageJ was used to exclude everything outside the folio of the manuscript. The MSA514 plugin was then run and was told how many images to produce (i.e. 5 in this case). An image stack is produced by the MSA514 plugin and by scrolling through the produced grayscale images it was seen that the undertext was mostly visible in channels 1, 4 and 5. The stack to RGB command was then used to produce a color image.

Each dimensionality reduction method produces a number of regression coefficients that are multiplied with the entire set of 23 multispectral images. This results in a new set of 23 arrays of floating point numbers which are further processed by rescaling the floating point numbers into the range of (0-255). This array will become a new 8 bit grayscale image.

A second set of 23 arrays of floating point numbers is produced based on the same multiplication between the regression coefficients and the entire set of multispectral images. The newly calculated minimum and the maximum values of the input points are scaled between 0 and 255 and the arrays of floating point numbers are rescaled based on this range. The scope of these two different processes is to map the new numbers to the range 0 and 255 by taking into account either the new scores of the input points or the new floating point numbers obtained from the above multiplication.

Finally, a third set of grayscale images can be obtained also for exploratory purposes by taking out the 0.01, 0.1, 1 or 5 percentiles of the data obtained from the multiplication of the regression coefficients with the entire set of multispectral images and following the same rescaling explained above. In this way it is possible this way to see the importance of the removed segments of the data.

The post-processing steps described above were applied identically to all the results obtained with the various dimensionality reduction methods.

All the grayscale images produced by the dimensionality reduction methods were investigated.

There were no pre-processing steps applied on the input data matrix, which consisted of the 23 rows by the 200 columns. However, some of the dimensionality reduction methods apply some pre-processing steps on the input data before applying the technique. Recentering the data on the mean and with variance 1 is applied in several techniques, such as NCA, GPLVM, LDA, t-SNE, PCA and DM. All the dimensionality reduction methods were used with the default input parameters. Most of these dimensionality reduction methods also calculate the eigenvectors in one of their inner computational steps, such as: CVA, PCA, DM, HLLE, t-SNE, GDA, GPLVM, LDA, NPE, Isomap, Landmark Isomap, LLE and Laplacian Eigenmaps.

---

[39] Pormann 2015.

Furthermore, a color image can be produced by combining three grayscale images. Normally, for each of the dimensionality reduction method, the grayscale image with the clearest underwriting was placed in the Green channel before producing the color image by combining Red, Green and Blue grayscale images. Further image enhancement can be achieved, for example, by adjusting the contrast on the grayscale images.

For the purposes of comparing the computational times with the CVA method implemented in Matlab, a CVA function was implemented in the C programming language using a software library for numerical computations called **GNU** (**G**NU's **N**ot **U**nix) **S**cientific **L**ibrary (**GSL**). For ease of use, ImageJ software[40] was employed, which is open architecture image processing software, which gives the possibility to add new functions/procedure/macros by writing Java plugins. By using a JNILIB library file, which is Java framework that allows Java to integrate with other programming languages, the CVA function implemented in C programming language as a JNILIB library file is called. The CVA-GNU GSL method is able to process and to produce both 8-bit and 16-bit images and the OpenCV[41] (OPEN source Computer Vision) computing programming library was used with this scope.

## 3. Evaluation of dimensionality reduction methods for ancient manuscripts

We used two approaches to evaluate the success of the image processing techniques. Firstly, the relative success of these methods was determined visually by seven experts in the Syriac language based on how well the scholars could read the undertext by distinguishing it from the parchment and the overtext. No further changes were made to the resulted images to directly assess the quality of the results of the dimensionality reduction methods. The scholars/experts were also able to identify the improvements the different dimensionality reduction methods achieved when compared to the original multispectral images.

Secondly we calculated two different indices that are commonly used for evaluating the success of multidimensional clustering techniques to see whether either of these agreed with the qualitative evaluations. The assessment made by the scholars is the standard way of evaluating these images, but for exploratory purposes the numerical comparison was also investigated (i.e. interferometric visibility another numerical method for quantitative assessment of manuscripts[20]). The first is the Davies-Bouldin Index (DBI)[42], which is one of the standard measures for evaluating clustering algorithms[43]. It is calculated using the following equations:

$$S_i = \frac{1}{T_i} \sum_{j=1}^{T_i} \left\| X_j - A_i \right\|_p \qquad (1)$$

where $S_i$ is a measure of scatter within the cluster i (i.e. the average distance between each point in the $i$ cluster and the centroid of the $i$ cluster), $T_i$ is the size of cluster, $A_i$ is the centroid of cluster $i$, $X_j$ are values forming a cluster and $p$ is usually 2.

Equation (2) describes $M_{i,j}$, the Euclidian distance between the centroids of the two clusters $i$ and $j$.

$$M_{i,j} = \left\| A_i - A_j \right\|_p = \left( \sum_{k=1}^{n} \left| a_{k,i} - a_{k,j} \right|^p \right)^{\frac{1}{p}} \qquad (2)$$

where in this case the two clusters are the underwriting cluster and the parchment cluster, $n$ is the size of the centroids $A_i$ and $A_j$, $a_{k,i}$ and $a_{k,j}$ are the $k$th element of clusters $A_i$ or $A_j$.

The measure of the effectiveness of clustering technique (i.e dimensionality reduction method) is given by $R_{i,j}$ where lower values, result from better the separation between the parchment cluster and the underwriting cluster:

---

[40] Schneider, Rasband and Eliceiri 2012.

[41] http://opencv.org

[42] Bouldin and Donald 1979.

[43] Franti, Rezaei and Zhao 2014.

$$R_{i,j} = \frac{S_i + S_j}{M_{i,j}} \tag{3}$$

where $S_j$ is a measure of scatter within the cluster j (i.e. the average distance between each point in the *j* cluster and the centroid of the *j* cluster).

For exploratory purposes a second well-known measure is used, which is known as the Dunn Index (DI)[44]. This index is suggested for clusters, with small variance between the different items of the clusters and with the mean values of different clusters being at a sufficiently large distance, which might be expected if our multispectral enhancement methods have performed well. Therefore DI is expected to prefer the CVA or LDA results, which are very likely to produce sets of clusters of the above type. However, the DI might be not so suitable for all the dimensionality reduction methods, hence the interest in exploring this index as well. The DI is defined by:

$$DI = \frac{Minimum\ Distance\ Between\ Cluster\ i\ and\ j}{Maximum\ Distance\ Within\ a\ Cluster\ Over\ All\ Clusters} \tag{4}$$

The minimum distance between cluster *i* and *j* is taken as the difference between $M_{i,j}$ and the scatters of the two clusters $S_i$ and $S_j$. Furthermore the maximum distance within a cluster and over all the clusters is taken as the maximum over the scatters of the two clusters $S_i$ and $S_j$, therefore:

$$DI = \frac{M_{ij} - S_i - S_j}{\max(S_i, S_j)} \tag{5}$$

Figure 1 shows a geometrical interpretation of the minimum distance between cluster *i* and *j*, which was taken as the difference between $M_{i,j}$ (i.e. the Euclidian difference between the means/centroids of the two clusters) and the scatters of the two clusters $S_i$ and $S_j$.

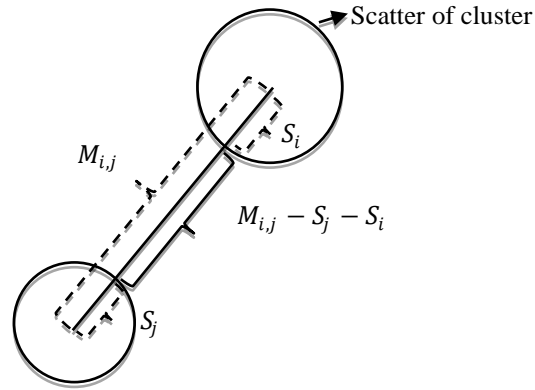

Figure 1. Geometrical interpretation of the minimum distance between cluster *i* and *j* taken as the difference $M_{i,j}$ (i.e. the Euclidian difference between the centroids of the two clusters) and the scatters (i.e. the average distance between each point in the cluster and the centroid of the respective cluster) of the two clusters $S_i$ and $S_j$.

## 4. Results

We chose 12 methods from the Matlab toolbox[30] and applied them to the 102v-107r_B page[45] of the Galen palimpsest and also applied the CVA method[39] used previously. In total, 13 supervised methods were investigated, from which 4 methods used both a number of user selected input points and the class information: Canonical Variates Analysis (CVA) method, Generalized Discriminant Analysis (GDA), Linear Discriminant Analysis (LDA) and Neighborhood Component Analysis (NCA). 6 supervised methods were also applied which used only the user selected input points: Gaussian Process Latent Variable Model (GPLVM), Isomap, Landmark Isomap, Principal Component Analysis (PCA), Probabilistic Principal Component Analysis (PPCA) and Diffusion Maps (DM). Another three supervised methods, which used only the user selected input points but did not give good results in this work, were Neighborhood Preserving Embedding (NPE) method, the t-Distributed Stochastic Neighbor Embedding (t-SNE) and the Hessian Locally-Linear Embedding (HLLE).

---

[44] Dunn 1973.

[45] URL: http://www.digitalgalen.net/Data/102v-107r/.

Finally, a regular image enhancement method known as the Double Thresholding (DT) method was applied and two independent implementations of the unsupervised PCA method in ImageJ and Matlab (i.e. making a total of 15 dimensionality reduction methods being used). The image results of the DT method vary function of the thresholds used.

In Figure 2 is shown an area of the color or grayscale images of the 13 supervised methods, the two unsupervised PCA methods (ImageJ, Matlab), the DT technique, the ultraviolet illumination with green color filter and the original page seen by the human eye for the 102v-107r_B page of the Galen palimpsest.

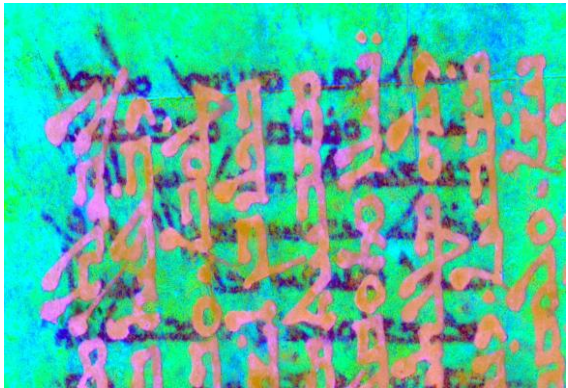
a) Canonical Variates Analysis method

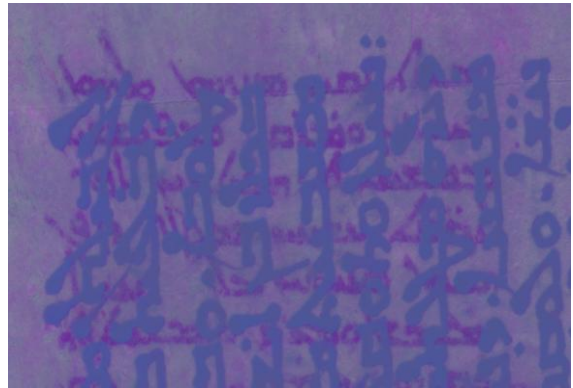
b) Linear Discriminant Analysis method

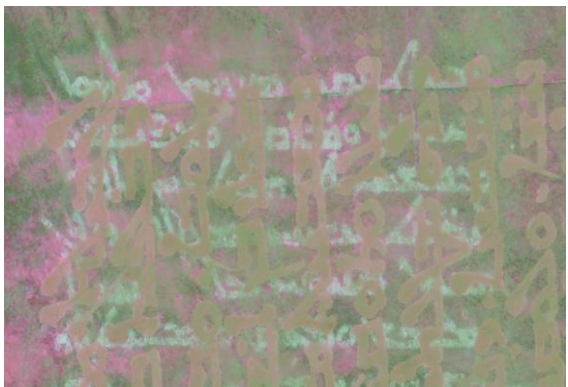
c) Neighborhood Component Analysis

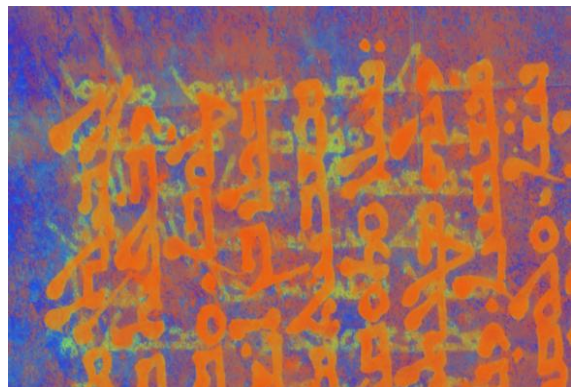
d) Generalized Discriminant Analysis

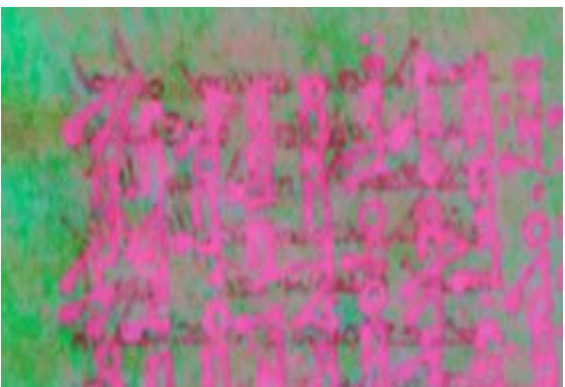
e) Diffusion Map

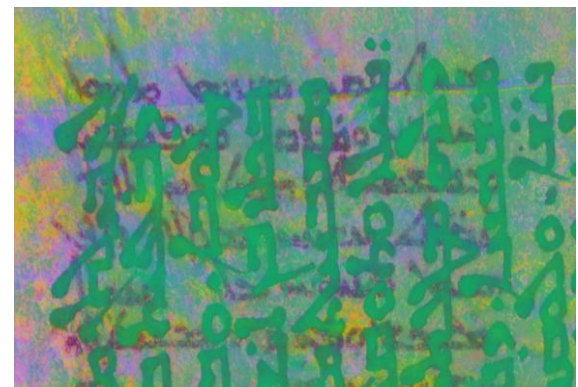
f) Isomap

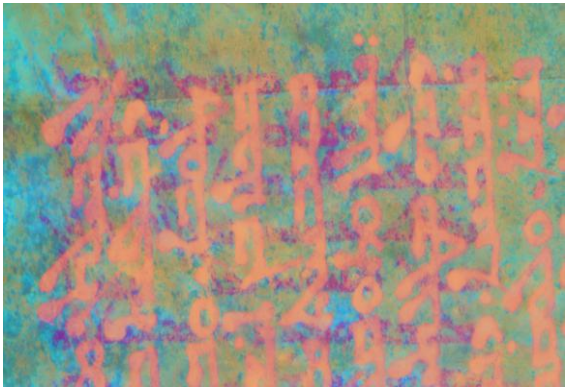

g) Landmark Isomap

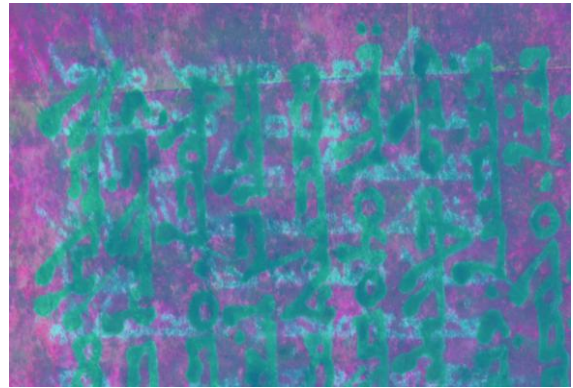

h) Principal Component Analysis-Unsupervised (ImageJ implementation)

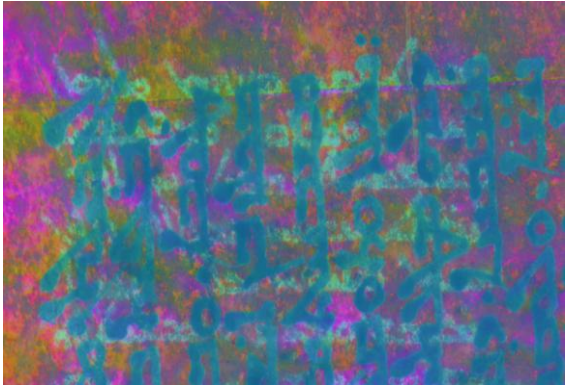

i) Principal Component Analysis

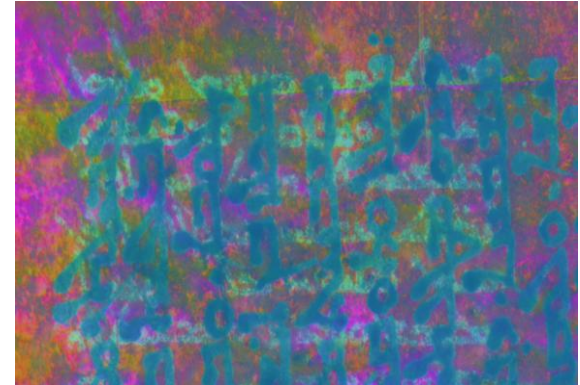

j) Gaussian Process Latent Variable Model

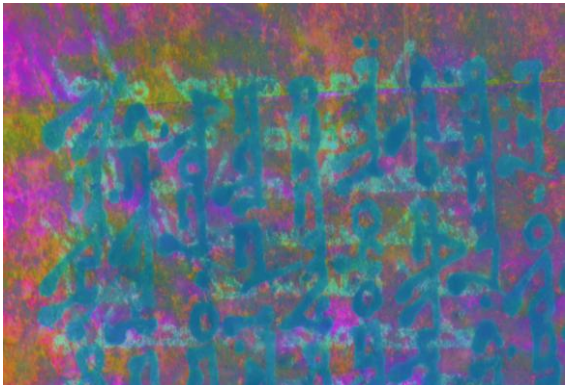

k) Probabilistic Principal Component Analysis

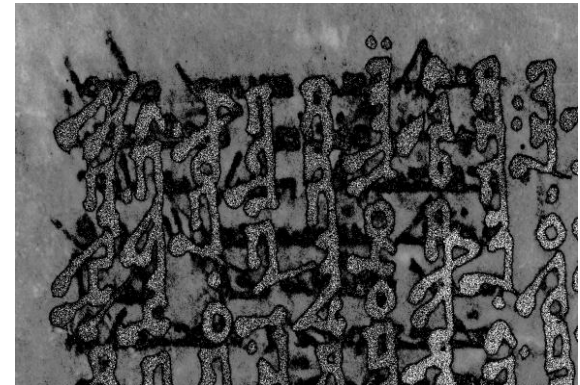

l) Double thresholding

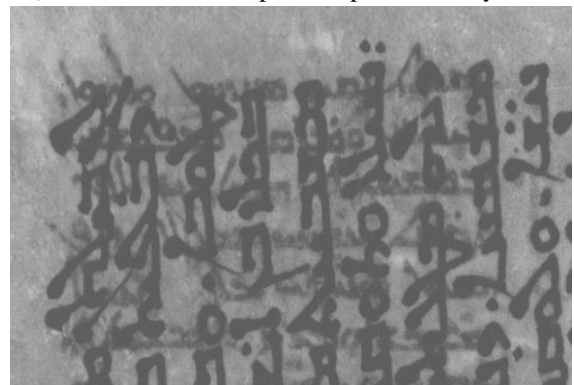

m) ultraviolet illumination with green color filter (i.e. CFUG)

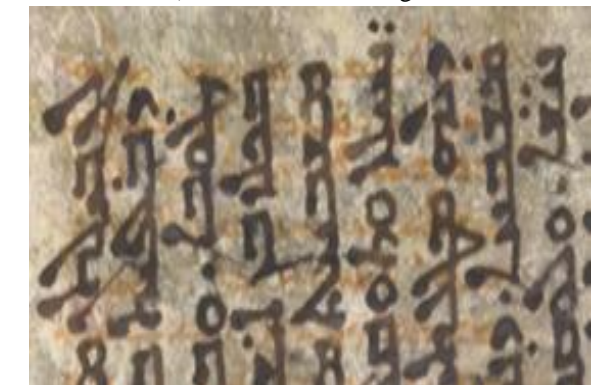

n) original page seen by the human eye

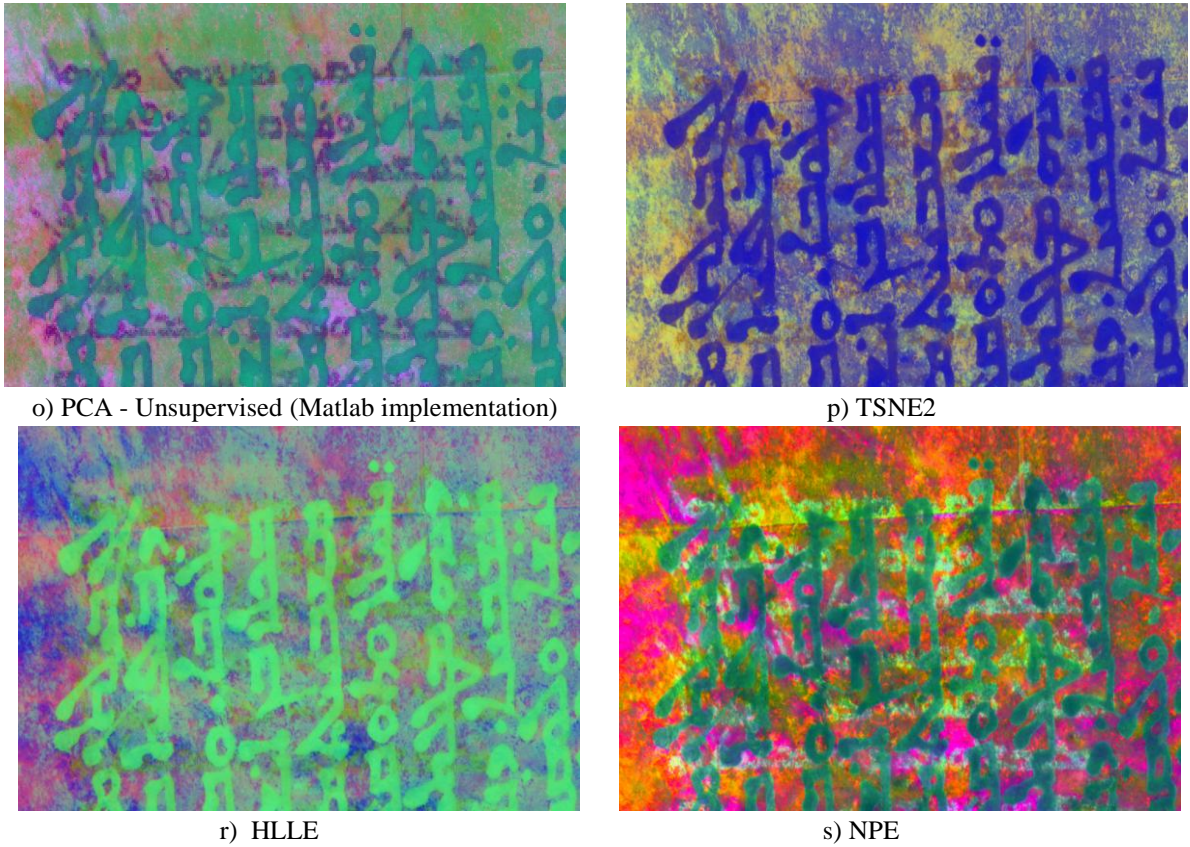

| | | |
|---|---|---|
| o) PCA - Unsupervised (Matlab implementation) | | p) TSNE2 |
| r) HLLE | | s) NPE |

Figure 2. Color image results obtained with 13 supervised dimensionality reduction methods, a simple double thresholding technique, two unsupervised PCA dimensionality reduction methods and in comparison with the original page seen by the human eye and the image obtained with the ultraviolet illumination with green color filter for a section of 102v-107r_B page.

As previously described, for the 13 supervised methods 50 points (i.e. 50 different x and y image pixel coordinates) were selected from each of the multispectral images for each of the classes Overwriting, Underwriting, Parchment and representing both overwriting and underwriting. 200 input points were used from each multispectral image, which resulted in an input matrix of 23 rows by 200 columns. This was used as the input data for each supervised dimensionality reduction method. In this case, the PCA, PPCA and GPLVM methods (Figure 2) gave similar visual and numerical results as can be seen above.

The visual assessment of the images in Figure 2 made by the 7 scholars experts in the Syriac language was done in two ways. First, by giving a score between 0 and 5 for the readability of the underwriting in each photo. The scores used were 5 – excellent, 4 – good, 3 – moderate, 2 – fair, 1 – poor, 0 – no readability and the images with the highest scores were deemed the best by the scholars in terms of underwriting. The total score was summed through the 7 lists produced by the 7 scholars. The first four images which scored best were CVA, PCA Matlab (unsupervised), GDA and Isomap (Table 1).

| | P1 | P2 | P3 | P4 | P5 | P6 | P7 | Total |
|---|---|---|---|---|---|---|---|---|
| CVA | 4 | 5 | 5 | 5 | 5 | 5 | 5 | 34 |
| PCA unsupervised (Matlab) | 4 | 4 | 5 | 2 | 5 | 4 | 4 | 28 |
| GDA | 4 | 4 | 3 | 3 | 5 | 3 | 4 | 26 |
| Isomap | 4 | 3 | 4 | 3 | 5 | 3 | 4 | 26 |
| LDA | 4 | 4 | 3 | 2 | 5 | 4 | 2 | 24 |
| PCA unsupervised (ImageJ) | 3 | 3 | 4 | 3 | 4 | 2 | 4 | 23 |
| NCA | 2 | 4 | 3 | 2 | 4 | 4 | 3 | 22 |
| DM | 2 | 3 | 3 | 2 | 5 | 2 | 4 | 21 |
| PCA | 1 | 3 | 4 | 2 | 4 | 2 | 3 | 19 |
| GPLVM | 1 | 3 | 4 | 2 | 4 | 2 | 3 | 19 |
| PPCA | 1 | 3 | 4 | 2 | 4 | 2 | 3 | 19 |
| Landmark Isomap | 3 | 3 | 3 | 2 | 3 | 1 | 4 | 19 |
| CFUG (ultraviolet) | 3 | 1 | 3 | 2 | 3 | 3 | 3 | 18 |
| Original page | 0 | 0 | 2 | 0 | 2 | 2 | 1 | 7 |
| DT | 0 | 1 | 1 | 1 | 0 | 1 | 2 | 6 |
| NPE | 0 | 0 | 1 | 1 | 1 | 2 | 1 | 6 |
| TSNE2 | 0 | 0 | 1 | 0 | 1 | 1 | 0 | 3 |
| HLLE | 0 | 0 | 0 | 0 | 0 | 0 | 0 | 0 |

Table 1. Scores between 5 and 0 given by the 7 scholars (P1-P7) experts in Syriac language
(5 – excellent, 4 – good, 3 – moderate, 2 – fair, 1 – poor, 0 – no readability).

Running an ANalysis Of VAriance (ANOVA) test for each column, where each column represents a different scholar, resulted in a p value of 0.0591. This result means that there are overall no significant differences between the different persons. Calculating the standard deviation of the Total column gives a value of 9.4903. We define the most effective methods as lying within one standard deviation of the top value, which is CVA (34), giving the following four methods: CVA, PCA Matlab (unsupervised), GDA and, Isomap.

| | P1 | P2 | P3 | P4 | P5 | P6 | P7 | Total |
|---|---|---|---|---|---|---|---|---|
| CVA | 1 | 1 | 1 | 1 | 1 | 1 | 1 | 7 |
| PCA unsupervised (Matlab) | 5 | 9 | 2 | 3 | 3 | 2 | 4 | 28 |
| NCA | 9 | 3 | 8 | 2 | 7 | 3 | 2 | 34 |
| Isomap | 4 | 6 | 3 | 4 | 4 | 5 | 8 | 34 |
| LDA | 2 | 2 | 10 | 9 | 2 | 4 | 6 | 35 |
| GDA | 3 | 4 | 9 | 5 | 6 | 7 | 3 | 37 |
| DM | 10 | 5 | 11 | 7 | 5 | 9 | 5 | 52 |
| PCA unsupervised (ImageJ) | 7 | 8 | 4 | 8 | 8 | 11 | 7 | 53 |
| Landmark Isomap | 6 | 7 | 12 | 6 | 12 | 8 | 9 | 60 |
| PCA | 11 | 10 | 5 | 10 | 9 | 12 | 10 | 67 |
| GPLVM | 12 | 11 | 6 | 11 | 10 | 13 | 11 | 74 |
| CFUG (ultraviolet) | 8 | 14 | 13 | 13 | 13 | 6 | 13 | 80 |
| PPCA | 13 | 12 | 7 | 12 | 11 | 14 | 12 | 81 |
| Original page | 16 | 15 | 14 | 14 | 15 | 10 | 14 | 98 |
| DT | 18 | 13 | 17 | 15 | 14 | 15 | 16 | 108 |
| TSNE2 | 14 | 16 | 15 | 16 | 17 | 17 | 17 | 112 |
| NPE | 15 | 18 | 16 | 18 | 16 | 16 | 15 | 114 |
| HLLE | 17 | 17 | 18 | 17 | 18 | 18 | 18 | 123 |

Table 2. Ranking positions (1 to 18) given by the 7 scholars (P1-P7) experts in Syriac language.

Finally, a second ANOVA test was performed on each row. The calculated p-value of 5.1596e-23 means that there are statistically significant differences between the scores given to the different images, which is obviously what was expected as some methods produced much better images than the others.

The second way in which the scholars assessed the images was to assign a rank from 1 to 18 based on how well the underwriting was readable and then to sum up the ranks for each image. The first six images which scored best (Table 2) corresponded to the methods CVA, PCA Matlab (unsupervised), NCA, Isomap, LDA, GDA. The standard deviation is 34.39 for the last column from Table 2 (Total) and the value of the best scoring method was added to this, yielding 41.39. The methods up to the value of 41.39 are in order CVA, PCA Matlab (unsupervised), NCA, Isomap, LDA and, GDA, which are exactly the same as the first six methods from above. It can also be observed that there are some strong similarities with the first four methods from Table 1.

For exploratory purposes, a numerical assessment of the color images from Figure 2 was done based on the grayscale images which had the clearest underwriting. A color RGB image is usually obtained by combining three of these grayscale images.

The grayscale image for which the underwriting is most visible after the application of a dimensionality reduction method, is, as already described, usually located in the green channel of the resultant color RGB image. 200 points from the underwriting cluster were compared with 200 points from the parchment cluster using the DB index, for the best looking grayscale image. Figure 3 shows the ranking of the grayscale images and the best looking grayscale image has the smallest value of the DB index.

| Method | Score | Method | Score |
|---|---|---|---|
| CVA | 0.0522 | PCA | 0.3 |
| Double Thresholding | 0.13 | PPCA | 0.3 |
| LDA | 0.2 | GPLVM | 0.3 |
| Ultraviolet illumination with green color filter | 0.21 | PCA unsupervised (ImageJ) | 0.331 |
| DM | 0.22 | PCA unsupervised (Matlab) | 0.38 |
| GDA | 0.235 | Original Image | 0.4194 |
| Isomap | 0.25 | TSNE2 | 0.6614 |
| Landmark Isomap | 0.283 | HLLE | 1.61 |
| NCA | 0.29 | NPE | 3.85 |

Table 3. DB index.

The DB index (Table 3) partly agreed with the visual assessment as the CVA method was the best (0.0522), followed in order by the Double Thresholding technique (0.13), LDA (0.2), the original 102v-107r_B page with ultraviolet illumination and green color filter (0.21), DM (0.22), GDA (0.235), Isomap (0.25), Landmark Isomap (0.283), NCA (0.29), PCA (0.3), PPCA (0.3), GPLVM (0.3), PCA unsupervised-ImageJ implementation (0.33), PCA unsupervised-Matlab implementation (0.38), Original page (0.4194), TSNE2 (0.6614), HLLE (1.61) and NPE (3.85). The CVA method gave a much better result (0.0522) than the 102v-107r_B folio with ultraviolet illumination with green color filter (0.21).

For exploratory purposes, the DI (Table 4), was also used to assess the respective grayscale images, which resulted in the rankings shown in Figure 4. The DI is suitable especially for CVA or LDA methods. Although the DI confirmed the superiority of the CVA method, overall the DB index was more conservative. The DI ranking of the methods was CVA (32.90), Double Thresholding (10.52), LDA (7.65), DM (6.45), NCA (4.78), GDA (4.39), Landmark Isomap (3.93), PCA (3.83), PPCA (3.83), GPLVM (3.83), Isomap (3.18), PCA unsupervised-Matlab implementation (3), PCA unsupervised-ImageJ implementation (2.94), Original image (2.46), TSNE2 (0.9963), HLLE (-0.69) and NPE (-1.191).

It has to be stressed again that although the indices are useful to be implemented for exploratory purposes, the standard way of assessing the resultant images is visually by scholars/experts in the respective language(s).

| Method | Score | Method | Score |
|---|---|---|---|
| CVA | 32.90 | GPLVM | 3.83 |
| Double Thresholding | 10.52 | PCA | 3.82 |
| LDA | 7.65 | Isomap | 3.18 |
| DM | 6.45 | PCA unsupervised (Matlab) | 2.94 |
| Ultraviolet illumination with green color filter | 6.30 | Original page | 2.46 |
| NCA | 4.78 | PCA unsupervised (ImageJ) | 2.26 |
| GDA | 4.39 | TSNE2 | 0.9963 |
| Landmark Isomap | 3.93 | HLLE | -0.69 |
| PPCA | 3.83 | NPE | -1.191 |

Table 4. Dunn index.

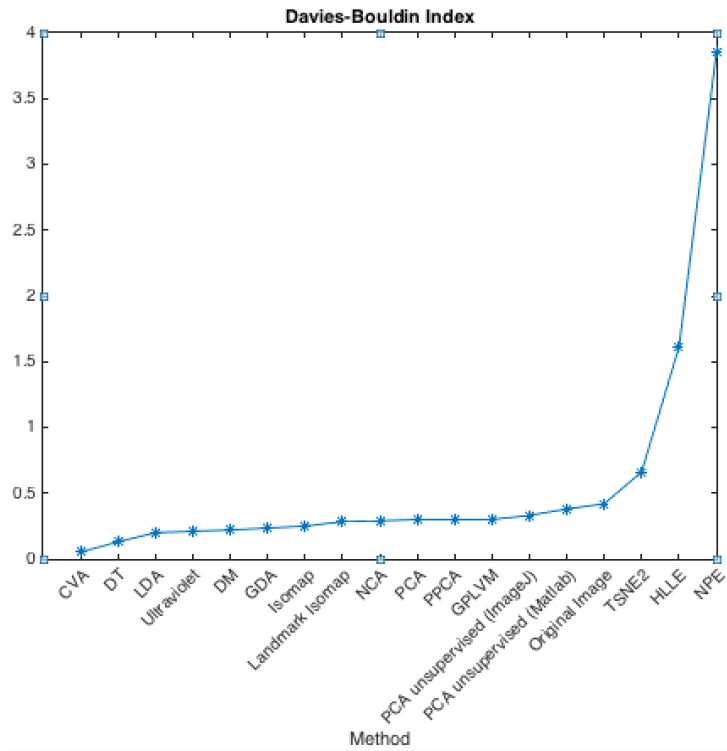

Figure 3. Numerical assessment of images based on Davies-Bouldin index.

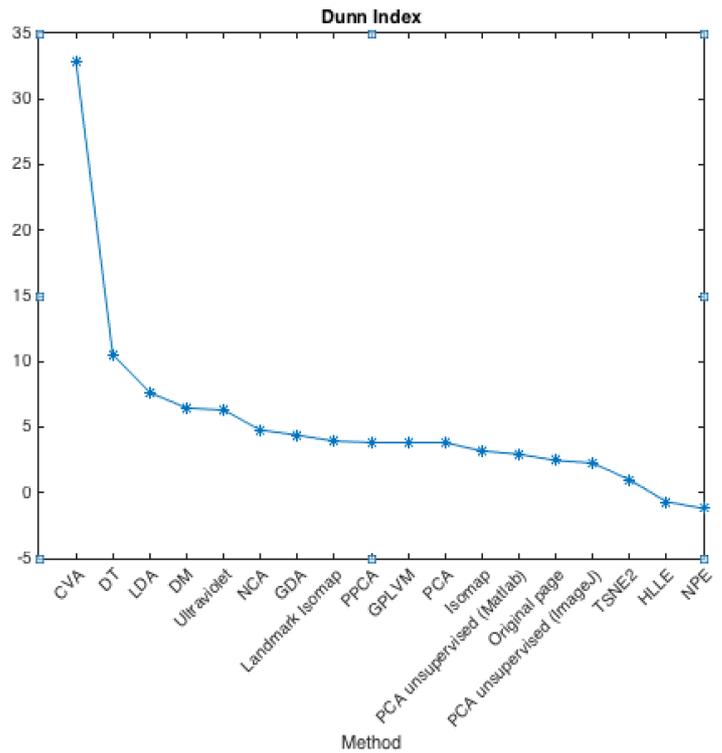

Figure 4. Numerical assessment of images based on Dunn index.

Comparing the numerical assessments based on DB and Dunn indexes versus the visual assessments, it is clear that there are a number of differences between the two assessments. The scholars did not find the DT method or the ultraviolet illumination with green color filter (i.e. CFUG) very useful, even though the numerical indexes indicated the two results were of better quality. This is because that, although the differences between the underwriting and parchment were significant compared with the underwriting, overall the images were not clear enough for the scholars. On the other hand, the PCA unsupervised method with the two different implementations in Matlab and ImageJ produced good images in terms of underwriting and were assessed positively by the scholars, even though the numerical indexes did not provide the same good results. This, suggests that, while the differences between the points representing the underwriting and the parchment were not significant in terms of the underwriting, there were not enough classification points being used (i.e. 200) in order to obtain important numerical results with the respective indexes.

Confidence intervals could be added to the results shown in Figure 3 and Figure 4 by applying the dimensionality reduction methods to a number of other palimpsest pages. However, the visual assessment done by the scholars was sufficient as the results were superior in terms of assessment quality than the numerical assessments with or without confidence intervals.

For the purposes of comparing the computational times and the image results, a CVA method is also implemented in C programming language using the *GNU* (*G*NU's *N*ot *U*nix) *S*cientific *L*ibrary (*GSL*) and OpenCV computer vision library. High performance software was developed by using the GNU GSL library, which drastically reduced the computational complexity and time for the CVA-GNU GSL method. When run on a 3.5 GHz Intel Core i5 with 16 GB RAM memory, the CVA-Matlab function together with the creation of the new color images (i.e. writing on the SSD hard drive), had a computational time of 111 seconds. On the same computer, the computational time for the CVA-GNU GSL was 80 seconds, which is 31 seconds shorter than the CVA-Maltab version, but this does not represent a critical time difference. This reduction of the computational time can be extended by using a computing programming library such as OpenMP[46] (OPEN MultiProcessing), which may further speed up the CVA-GNU GSL software by parallelizing the software code.

The eigen analysis functions in Matlab, *eig*, and GNU GSL, *gsl_eigen_gensymmv*, result in different calculated eigenvectors, for example, there can be different signs for the significant eigenvectors. This can be seen in Figure 5. Figure 5a is produced by the CVA-Matlab and Figures 5b and 5c are produced by CVA-GNU GSL. A similar result to Figure 5a was obtained by inverting the first two channels of the color image from the CVA-GNU GSL method (i.e. Figure 5b) and which is shown in Figure 5c. The majority of scholars (i.e. three out of five) considered the CVA-GNU GSL result (Figure 5b) better in terms of underwriting than the CVA-Matlab (Figure 5a), however, one scholar considered all the three results were of comparable high quality.

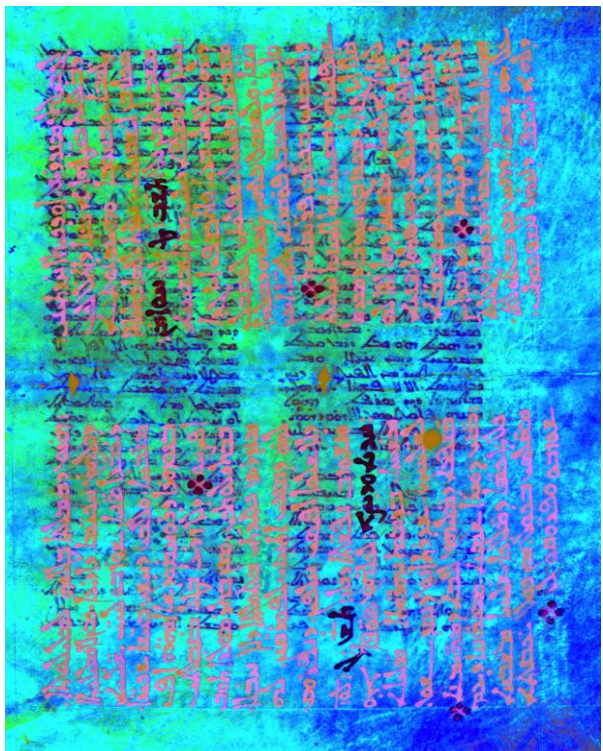 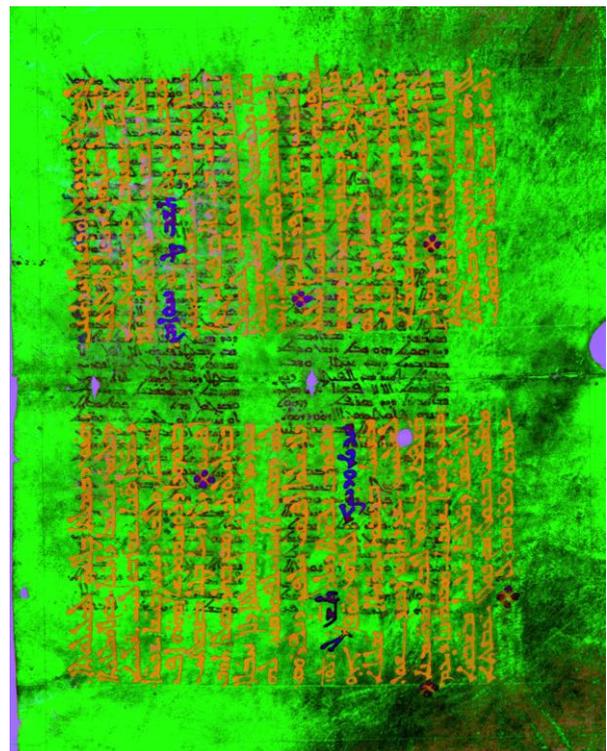

a) CVA-Matlab  b) CVA-GNU GSL

---

[46] http://www.openmp.org

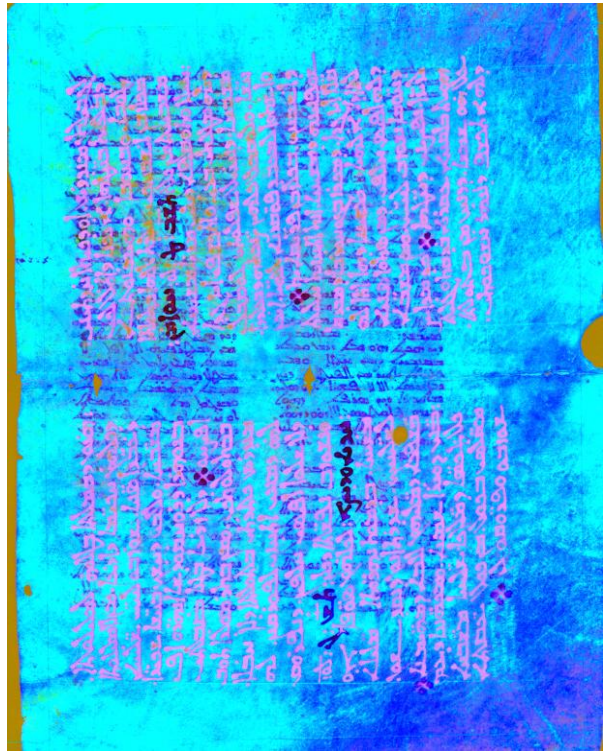

c) Inverting the first two channels of the three channel image (Red, Green and Blue) of the CVA-GNU GSL image result produces a photo similar to the CVA-Matlab result.

Figure 5. Color image results obtained with CVA-Matlab and CVA-GNU GSL.

The entire Syriac Galen palimpsest, which consists of about 240 pages, was processed with the CVA-GNU GSL, in less than 2 months and was able to produce both 8-bit and 16-bit images using the OpenCV computer vision programming library. Some pages were processed several times by varying the number of training input points to search for improvements in terms of the underwriting.

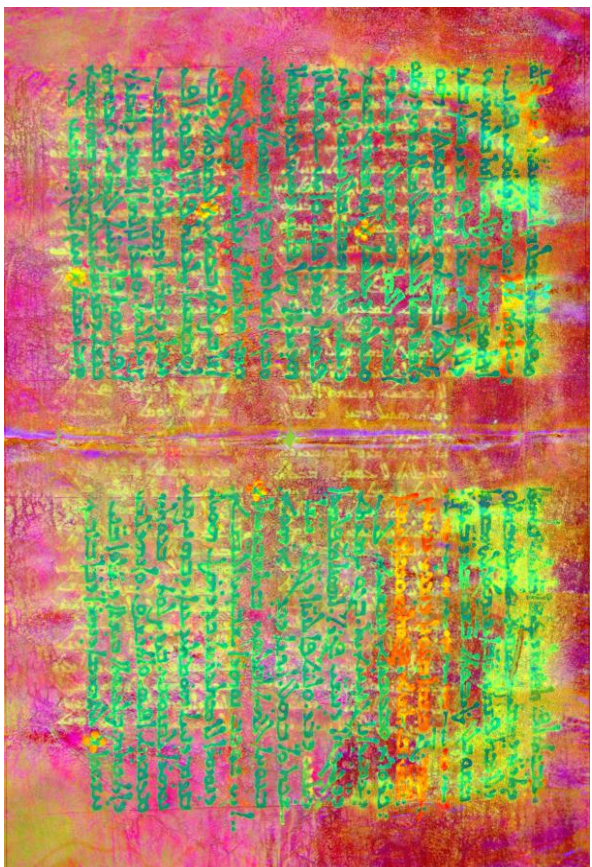

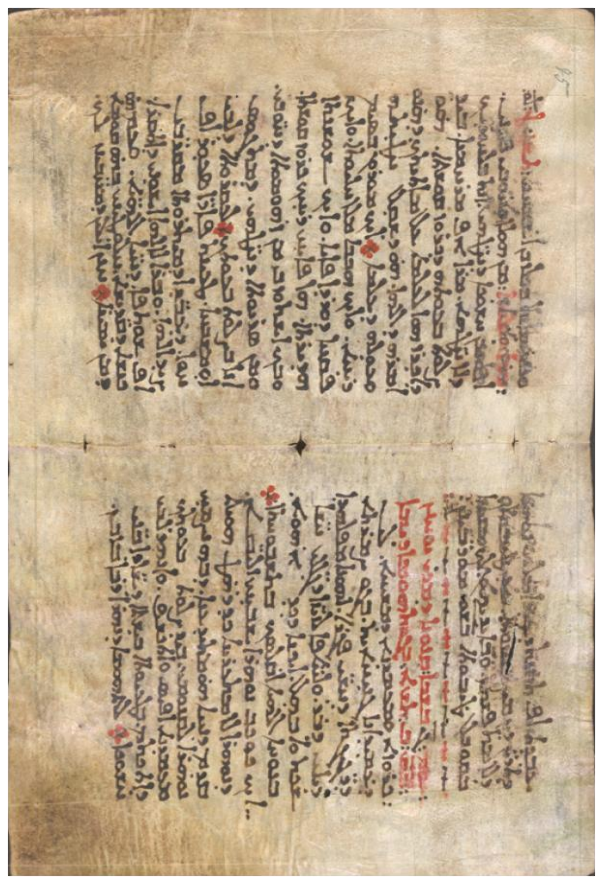

a) The CVA-GNU GSL processed page

b) Original page seen by the human eye

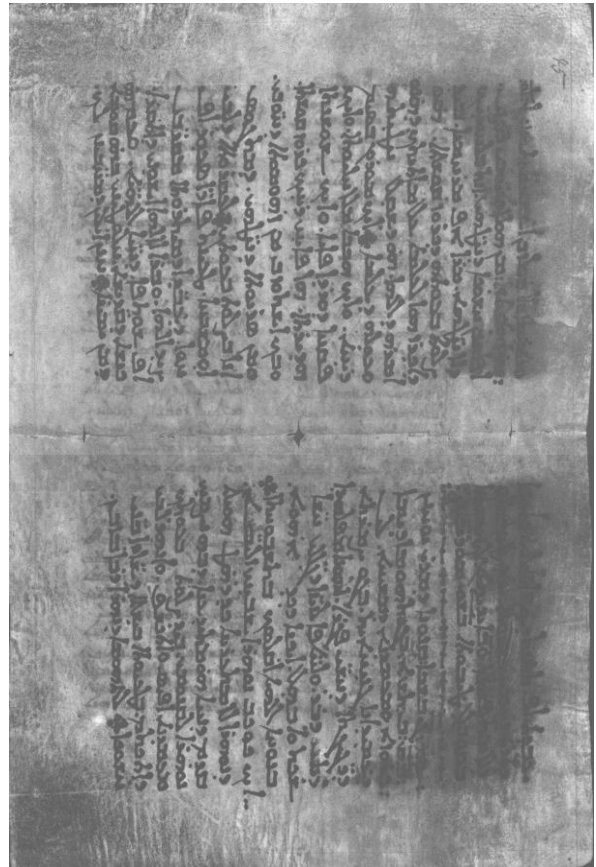

c) Ultraviolet illumination with green color filter

Figure 6. Comparison for folio 040v-045r between the CVA-GNU GSL processed page and the multispectral and the original page seen by the human eye.

Figure 6 shows another folio of the palimpsest processed with the CVA-GNU GSL software library and the original page seen by the human eye for comparison (i.e. folio 040v-045r). This result shows how the underwriting is revealed to the scholars after multispectral imaging and the application of CVA-GNU GSL method. In Figure 6 there is no underwriting in the original page that can be seen by the human eye, while in the multispectral image, acquired with ultraviolet illumination with the green color filter, the underwriting starts to be revealed. In the image result obtained with the CVA-GNU GSL the underwriting becomes quite clear all over the respective page.

The study carried out on the Galen palimpsest identified a smaller set of dimensionality reduction methods which provided the best image results. From this smaller set of methods, based on the visual and the numerical evaluation of the produced images, the CVA method produced the best result. Therefore we wanted to test the CVA on a second manuscript written in Latin language and entitled *John Rylands Library, University of Manchester, Latin MS 18*. The authorship and provenance are currently unknown so hence there is a high degree of interest in identifying the respective manuscript. The manuscript does not have any underwriting, so it is not a palimpsest, but some of the text has been almost deleted because of passing of time and water.

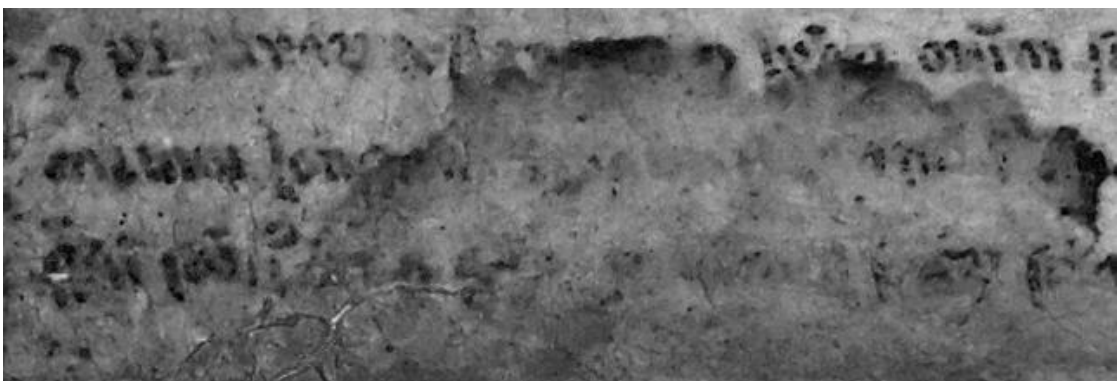

a) Section obtained with CVA-Matlab for a grayscale image.

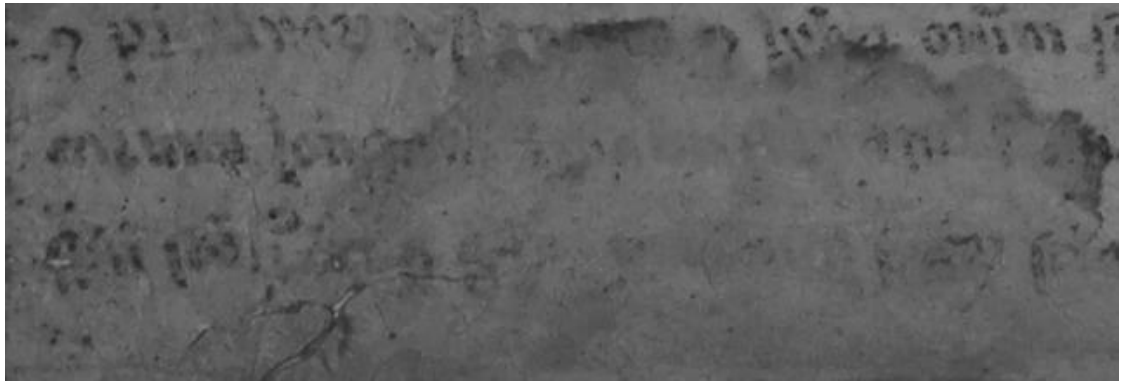
b) Section obtained from the multispectral image of the Latin MS 18 manuscript.
Figure 7. Comparison between a section of one of the grayscale images produced by the CVA-Matlab method and the same section obtained from one of the multispectral images.

In Figure 7 a result is shown, which was processed with CVA-Matlab with 200 points for the classes manuscript/parchment and text, for comparison with the multispectral image. For some sections of the page a slight improvement can be noticed with the CVA method, such as in Figure 7a.

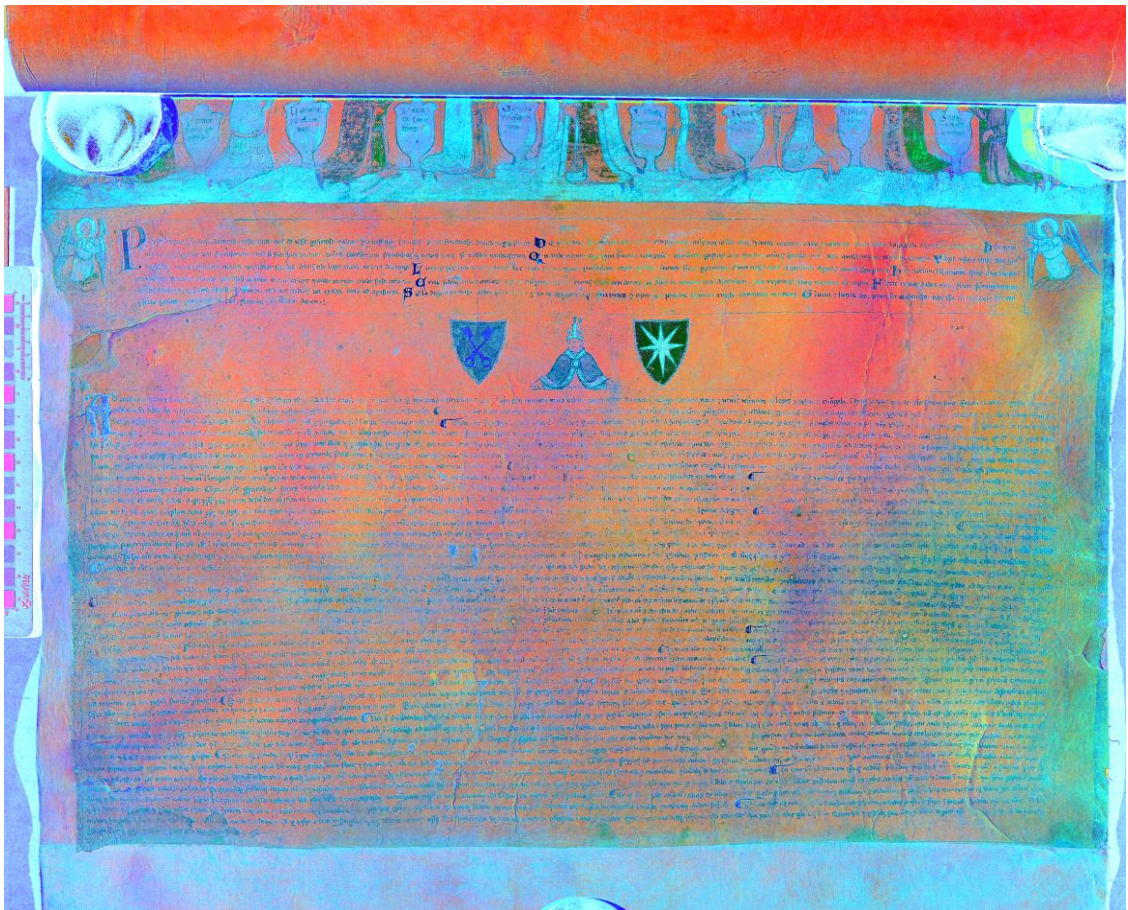
a) Color image obtained with CVA-Matlab.

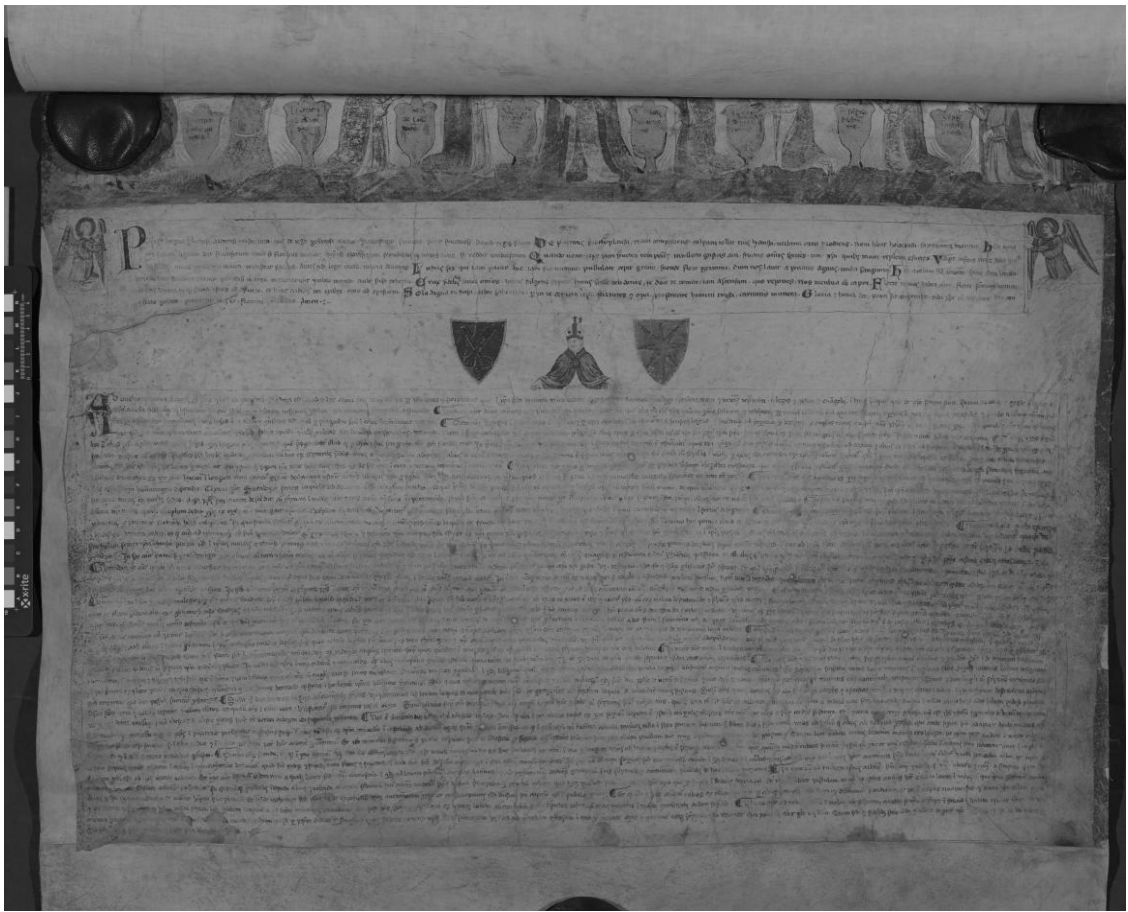

b) Multispectral image acquired for the Latin MS 18 manuscript.
Figure 8. Comparison between color image obtained with the CVA-Matlab and one of the multispectral images acquired for the Latin MS 18 manuscript.

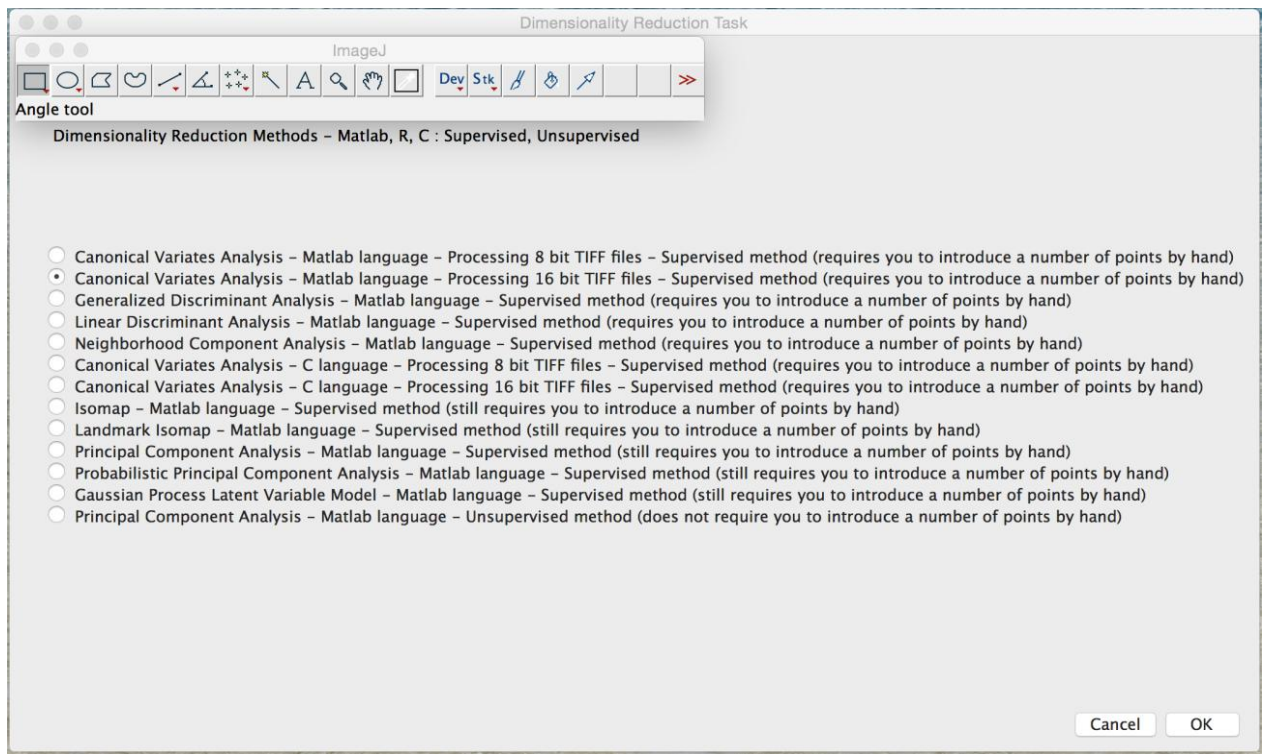

Figure 9. Screen capture with the plugin developed for the ImageJ software with which the various dimensionality reduction methods implemented in Matlab or C-GNU GSL can be called for the purposes of image processing.

In Figure 8, a comparison can be seen between the color image obtained with the CVA-Matlab and one of the multispectral images acquired for the Latin MS 18 manuscript.

In Figure 9 a screen capture can be seen with the plugin[47], developed for the ImageJ software, with which the various dimensionality reduction methods implemented in Matlab or C-GNU GSL can be called[48]. The methods can be run with various numbers of input points and classes (e.g. class underwriting, class overwriting, etc). The methods that are available at the present time in the plugin and implemented in Matlab are CVA (i.e. processing 8 and 16 bit images), LDA, DM, GDA, Isomap, Landmark Isomap, NCA, PCA, PPCA, GPLVM and the CVA-GNU GSL method implemented in C programming language is also available (i.e. processing 8 and 16 bit images).

The image results and their visual and numerical assessment confirmed that the methods CVA and LDA seem to be always in the leadership position with regard to the image quality. The CVA/LDA methods find the linear combinations of variables that maximize the separation of data classes which were defined a priori. This means that, given a set of input points representing the different data classes (i.e. class parchment, class overwriting, class underwriting), the CVA method should be able to produce images which provide the best visual distinction between the different classes representing parchment, overwriting and underwriting. This characteristic of the CVA method, which is distinct from the other dimensionality reduction methods being tested here, was the most successful method studied and should be used further.

The use of supervised methods may involve the additional step of selecting the input points, but, as already envisaged in the previous sections, an automated technique for choosing the input points and classes to which the respective points belong to, has been recently devised[49]. Further work will be to include a quick automated process for choosing the input points and the classes to be used with the CVA method, and to further speed up the processing of multispectral images.

Moreover, it was noted that when the unsupervised PCA method (i.e. Matlab, ImageJ) was used without selecting any input points, the image results obtained by PCA provided very good and important results in terms of underwriting. However, it must be stressed again that there is a need for an automated process to choose the input points and the classes to be used with supervised or unsupervised methods. More recently[49], another method was devised by which the overtext and the parchment were masked for a small section of a page of a palimpsest (i.e. a tile) and only the underwriting was used to train the PCA method. This last method resulted in very good image results and could also be considered for other supervised or unsupervised training methods.

Some other methods proved to give good results such as Isomap and Landmark Isomap, which confirmed similar findings from literature[34]. Hence there is an interest to investigate other dimensionality reduction methods as sometimes they could provide good image results.

Other image enhancement methods could be applied to the grayscale images obtained from using the dimensionality reduction methods. One way would be to further improve the contrast via linear or polynomial scaling, as discussed previously. Another technique can be implemented, similar to the pseudocolor technique described above, where the grayscale image with the clearest underwriting is used in two channels rather than one to improve the appearance of the resulted color RGB image, which is usually obtained by combining three different grayscale channel images.

Finally, several dozen more images produced with the CVA method in Matlab or GNU GSL for the *Syriac Galen Palimpsest* showed at least the same quality, if not better, when compared with images produced by other image processing methods[50], in most of the cases. The entire *Syriac Galen Palimpsest*[51] was processed with the CVA-GNU GSL method and OpenCV library together with the developed ImageJ plugin, which consists of a number of Java files. These files are called by ImageJ software and also call the developed software library implementing the CVA-GNU GSL method. More than a quarter (i.e. 68 folios) of the *Syriac Galen Palimpsest* processed with the CVA-GNU GSL method is freely available online[52] for any scholar to download and study.

---

[47] https://zenodo.org/record/154127#.WVD12WgrJEY

[48] https://github.com/corneliu25/GalenProject

[49] Hollaus et al. 2015.

[50] Easton et al. 2010; Bergmann and Knox 2009; Hollaus et al. 2015.

[51] Afif et al. 2017; Afif et al. 2016.

[52] https://zenodo.org/record/252293#.WVEeaGgrJEY

## 5. Conclusions

Our findings suggest that a supervised dimensional reduction technique, such as CVA, is an excellent processing tool for multispectral images. The choice of method is ultimately based on the preferences of the person trying to read the manuscript, the precise makeup of the original document and obviously the quality of the images produced by the respective dimensionality reduction method. In addition to these, easy access to an appropriate toolset software was clearly highly desirable to provide quick processing of the multispectral images. The use of existing software for image processing, ImageJ, and the addition of the extra functionalities in course of this work, provided fast and remarkably good processing of the multispectral images of the Galen palimpsest, of the second unidentified Latin manuscript and a number of other manuscripts.

Further work will consist of applying these dimensionality techniques and the software developed to enable the recovery of the undertext or hard-to-read texts in various other palimpsests and ancient manuscripts.

Acknowledgment
The authors would like to thank the Arts and Humanities Research Council, United Kingdom, for supporting this work (Research Grant AH/M005704/1 - The Syriac Galen Palimpsest: Galen's *On Simple Drugs* and the Recovery of Lost Texts through Sophisticated Imaging Techniques). The authors would like to thank to Prof Peter E Pormann, Dr Naima Afif, Dr. William Sellers, Dr. Natalia Smelova, Dr Siam Bhayro, Dr Taro Mimura, Dr Kamran Karimullah, Dr Grigory Kessel, Dr. Elaine Van Dalen, Dr Irene O'Daly, Dr. Michelle Magin, Dr. William Christens-Barry, Mr Michael Toth, Dr Chloe Jeffries, Mrs Gwen Riley Jones and Mrs Carol Burrows in supporting this work.